\pdfoutput=1

\documentclass[11pt]{article}

\usepackage[preprint]{acl}

\usepackage{times}
\usepackage{latexsym}

\usepackage[T1]{fontenc}

\usepackage[utf8]{inputenc}

\usepackage{microtype}

\usepackage{inconsolata}

\usepackage{graphicx}

\usepackage{amsmath}
\usepackage{booktabs}
\usepackage{cleveref}
\usepackage{makecell}
\usepackage{linguex}

\usepackage{etoolbox}
\usepackage{setspace}
\usepackage{subfig}
\AtBeginEnvironment{quote}{\singlespacing\small}

\title{One fish, two fish, but not the whole sea:\\ Alignment reduces language models' conceptual diversity}

\author{
 \textbf{Sonia K. Murthy\textsuperscript{1,2}},
 \textbf{Tomer Ullman\textsuperscript{2,3}},
 \textbf{Jennifer Hu\textsuperscript{2}}
\\
\\
 \textsuperscript{1}School of Engineering and Applied Sciences, Harvard University \\
 \textsuperscript{2}Kempner Institute for the Study of Natural and Artificial Intelligence, Harvard University \\
 \textsuperscript{3}Department of Psychology, Harvard University \\ 
 \texttt{soniamurthy@g.harvard.edu,\{tomerullman, jenniferhu\}@fas.harvard.edu}
}

\begin{document}
\maketitle
\begin{abstract}
Researchers in social science and psychology have recently proposed using large language models (LLMs) as replacements for humans in behavioral research. 
In addition to arguments about whether LLMs accurately capture population-level patterns, this has raised questions about whether LLMs capture human-like conceptual diversity. Separately, it is debated whether post-training alignment (RLHF or RLAIF) affects models' internal diversity.
Inspired by human studies, we use a new way of measuring the conceptual diversity of synthetically-generated LLM ``populations'' by relating the internal variability of simulated individuals to the population-level variability. We use this approach to evaluate non-aligned and aligned LLMs on two domains with rich human behavioral data. 
While no model reaches human-like diversity, 
aligned models generally display less diversity than their instruction fine-tuned counterparts. 
Our findings highlight potential trade-offs between increasing models' value alignment and decreasing the diversity of their conceptual representations. 
\end{abstract}

\section{Introduction}

As large language models (LLMs) have become more sophisticated, there has been growing interest in using them to replace human labor. 
This appeal of LLMs has even made its way to settings where \emph{human behavior itself} is the object of inquiry: recently, researchers have proposed that LLM-generated responses can be used in place of human data for tasks such as polling, user studies, and behavioral experiments \citep[e.g.,][]{aher_using_2023,argyle_out_2023,synthetic_hci_2023,binz_turning_2024,manning_automated_2024}. If possible, using a synthetic replacement for the process of human data collection
could be transformative for a variety of human factors disciplines, from political science to economics and psychology.

\begin{figure}[ht]
  \centering
  \includegraphics[width=\linewidth]{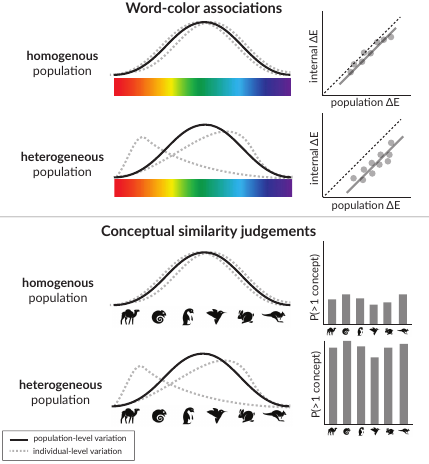}
  \caption{We investigate LLM populations comprised of simulated individuals in two domains: color associations (top) and concept similarity (bottom). In both domains, there is both individual- and population-level variation.
  It is possible that individual variation overlaps with the population average (homogeneous population) or separates from it (heterogeneous population). Our experiments are designed to tease these two options apart.}
  \label{fig:overview}
\end{figure}

While prior work suggests that LLMs can capture certain behavioral patterns, there are ongoing debates as to whether they are valid replacements for human subjects \cite{dillion_can_2023, wang_large_2024, park_diminished_2024}.
One key issue is whether LLMs capture \textbf{conceptual diversity}: the variation among \textit{individuals}' representations of a particular domain.
A natural way to study conceptual diversity is to study the variability in LLMs' response distributions at the \emph{population} level -- that is, by considering  averages across simulated individuals.
However, population-level variability can be a flawed measure of conceptual diversity for several reasons. First, it assumes certain characteristics about the nature of people's mental representations that may not hold in practice -- for example, that people have high certainty in these representations \citep[e.g.,][]{marti_certainty_2018}, or that people's responses are a deterministic best-guess, with no additional information contained in successive guesses \citep[e.g.,][]{Vul2008MeasuringTC}. 
Second, assessing population-level variation without individual-level variation can mask important information about the population, particularly whether it is \textit{homogeneous} (comprised of individuals who share similar underlying representations) or \textit{heterogeneous} (comprised of individuals with meaningfully different representations) (see \Cref{fig:overview}). 

Conceptual diversity may be negatively impacted by \textbf{alignment}, which further complicates the picture. Post-training alignment techniques, such as RLHF \citep{ouyang_training_2022,bai_training_2022} and RLAIF \citep{bai_constitutional_2022}, are now standard parts of LLM development, and presumed to contribute to the human-like abilities of models \citep{ji_ai_2024}. However, it has also been observed that ``aligned'' models show biases 
that can limit the lexical and content diversity of their outputs \citep{janusModeCollapse, padmakumar_does_2024, park_diminished_2024,omahony_attributing_2024}. It also remains unknown whether alignment to synthetic preferences (instead of human preferences) might worsen these biases, as it is possible these models ``collapse'' when recursively trained on synthetically generated data \citep{shumailov_ai_2024,gerstgrasser_is_2024}. 

In this paper, we (1) investigate the ability of LLMs to capture human-like conceptual diversity, and (2) analyze the effect of alignment by comparing conceptual diversity across non-aligned models to models aligned using RLHF or RLAIF.
In our experiments, we first simulate populations of unique individuals in LLMs using two techniques proposed by prior literature: temperature- and prompt-based manipulations. 
We then evaluate ten open-source LLMs on two domains with rich human behavioral data: word-color associations, and conceptual similarity judgments.
While there is no single agreed-upon metric for capturing conceptual diversity, we consider two different metrics that are each well-suited to their respective domains, to measure the conceptual diversity of synthetically-generated LLM ``populations''. 

We find that no model approaches human-like conceptual diversity. Further, aligned models generally display less conceptual diversity than their non-aligned, fine-tuned counterparts. 
Our findings suggest that there may be trade-offs between increasing model safety in terms of value alignment, and decreasing other notions of safety, such as the diversity of thought and opinion that models represent. We caution that these trade-offs should be better understood before models are used as replacements for human subjects, or deployed in human-centered downstream applications.

\begin{table*}[t]
    \footnotesize
    \centering
    \newcommand{\base}[1]{{\color{black!50}{#1}}}
    \newcommand{\hf}[1]{\texttt{#1}}
    \newcommand{\model}[2]{\makecell{#1 \\ {\scriptsize{(\hf{#2})}}}}
    \begin{tabular}{llll} \toprule
        \base{Base model} & Non-aligned variant & Aligned variant & Alignment method \\ \midrule
        
        \base{\model{Mistral}{mistralai/Mistral-7B-v0.1}} & \model{Openchat}{openchat/openchat\_3.5} & \model{Starling}{berkeley-nest/Starling-LM-7B-alpha} & RLAIF (APA) \\[1.8em]
        
        \base{\model{Mistral}{mistralai/Mistral-7B-v0.1}} & \model{Mistral-Instruct}{mistralai/Mistral-7B-Instruct-v0.1} & \model{Zephyr-Mistral}{HuggingFaceH4/zephyr-7b-beta} & RLAIF (DPO) \\[1.8em]
        
        \base{\model{Gemma}{google/gemma-7b}} & \model{Gemma-Instruct}{google/gemma-7b-it} & \model{Zephyr-Gemma}{HuggingFaceH4/zephyr-7b-gemma-v0.1} & RLAIF (DPO) \\[1.8em] 
        
        \base{\model{Llama}{meta-llama/Llama-2-7b-hf}} & \model{Llama}{meta-llama/Llama-2-7b-hf} & \model{Llama-Chat}{meta-llama/Llama-2-7b-chat-hf} & RLHF (PPO) \\[1.8em]
        
        \base{\model{Llama}{meta-llama/Llama-2-7b-hf}} & \model{Tulu}{allenai/tulu-2-7b} & \model{Tulu-DPO}{allenai/tulu-2-dpo-7b} & RLAIF (DPO) \\ \bottomrule
        
    \end{tabular}
    \caption{Pairs of non-aligned and aligned models tested in our experiments. Huggingface identifiers are shown in parentheses underneath model names. 
    The base models are shown for reference and are not directly tested, with the exception of the comparison between Llama and Llama-Chat. All models have 7B parameters.}
    \label{tab:models}
\end{table*}

\section{Background}

Recent works have proposed using LLMs as stand-ins for humans in many applications and domains, including opinion surveys and polling in political science \cite{argyle_out_2023}, user studies in human-computer interaction \cite{synthetic_hci_2023}, annotation tasks \cite{Gilardi2023ChatGPTOC}, and various economic, psycholinguistic, and social psychology experiments \cite{aher_using_2023, dillion_can_2023}.

Several works have responded by identifying LLMs' lack of answer diversity as a potential harm of using LLMs to replace human subjects. These works have focused on whether models are capable of capturing demographic- and subgroup-level variation in settings where accurately simulating the social identities of a human population of interest is directly relevant to the task. \citet{wang_large_2024} find that inference-time interventions to improve models' output diversity do not prevent LLMs from misportraying and flattening the representations of already marginalized demographic groups. Even after explicit steering toward certain demographics, LLMs still do not capture these group's responses on public opinion surveys \citep{santurkar_whose_2023}.

These observations have motivated recent attempts to go from broad, population-level observations of diversity, to metrics that assess finer-grained notions. 
For example, \citet{he-yueya_psychometric_2024} measured the extent to which LMs capture human knowledge distributions on real-word domains like language learning and mathematics.
Their metric captures whether two populations are similarly sensitive to the relative difficulty of test items. However, this metric is not well-suited for domains that lack ground-truth answers or where people's mental representations are not easily parameterized.
\citet{franke_bayesian_2024} propose another item-level account of diversity, from the perspective of Bayesian statistical modeling. They find that LLMs do not capture human variance at the item-level in an experiment on pragmatic language, and
aggregate item-level predictions to replicate condition-level effects seen in human data. Similarly, \citet{wang_not_2024} also find that LMs replicate broad-level patterns (e.g., mean and standard deviation) but fail to capture item-level patterns of human behavior. 
These studies highlight the importance of finer-grained comparisons between model and human outputs, but leave open the question of how to measure the relationship between individual-level variation and population-level variation in rich conceptual domains.

\begin{figure*}[h]
  \centering
  \includegraphics[width=\textwidth]{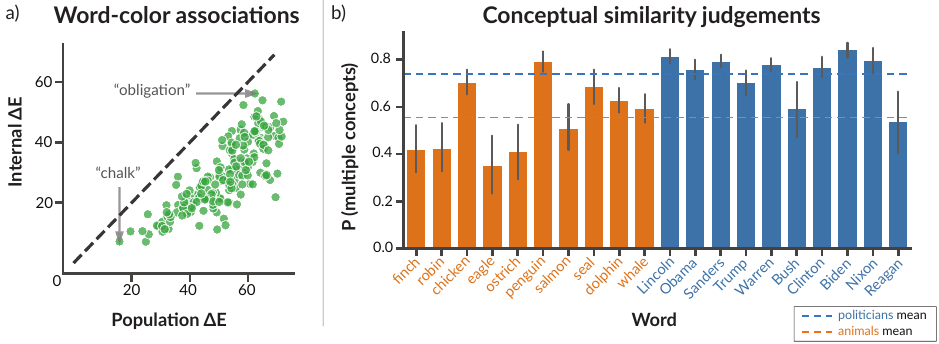}
  \caption{Human baselines in both domains. (a) Internal variability (y-axis) versus population variability (x-axis) for human participants in the word-color association domain. Reproduced from \citet{murthy_shades_2022}. The measures are correlated but clustered under the identity line (the internal variability is lower than population variability), indicating a heterogeneous population in terms of conceptual diversity. (b) Probability of more than one conceptual representation, estimated using Chinese Restaurant Process model on human data for conceptual similarity judgement task \citep{marti_latent_2023}.}
  \label{fig:baselines}
\end{figure*}

\section{Methods}

Our experiments are designed to investigate two related research questions: (1) Do modern LLMs capture the conceptual diversity of human populations, i.e., \textbf{individuals'} variability in conceptual representations? and (2) How does post-training alignment (here, RLHF or RLAIF) affect models' ability to capture this feature of response distributions from human populations?

The basic logic of our experiments is as follows. First, we identify domains where prior work has demonstrated rich conceptual diversity in humans. Second, we test whether LLMs capture conceptual diversity patterns by simulating ``populations'', using temperature- and prompting-based manipulations. Finally, we compare the conceptual diversity achieved by models against that of humans, and specifically examine the effect of alignment on the degree of fit between models and humans.

\subsection{Domain 1: Word-color associations} \label{sec:colors-domain}

The first domain probes people's intuitive associations between words and colors, building on the experiments of \citet{murthy_shades_2022}. 
In this task, human participants were presented with a target word (such as ``chalk'' or ``obligation'') and asked to click on the color most associated with the word, from a set of 88 color chips. Participants were tested in two blocks, ultimately providing two color associations for each word. This design enables measures of \emph{population} variability (how much color associations vary across individuals) as well as \emph{internal} variability (how much they vary in an individual's own representation). 

\paragraph{Model evaluation.}

To measure which color a model ``associates'' with a particular target word, we used the following query:\footnote{For the experiments that simulate unique subjects through prompting manipulations, the relevant context was prepended to this query (see \Cref{manipulations} for more details).}
\begin{quote}
    Question: What is the HEX code of the color you most associate with the word \texttt{[WORD]}? You must respond with a guess, even if you're unsure. Make sure your response contains a valid 6-digit HEX code.
\end{quote}
Conditioned on this input, models were allowed to freely generate subsequent text.\footnote{While the human baseline was collected through interaction with a visual color space, we adopt this task textual format following previous work \cite{Marjieh2023LargeLM, niedermann_studying_2024}}
Two samples were elicited for each word and HEX codes were extracted from models' generations using a regular expression.\footnote{When one or both samples contained no valid HEX code, the response was considered invalid and was excluded from analyses (see Appendix, \Cref{fig:response_counts} for the number of valid responses for each model and manipulation).}
This procedure was repeated to collect models' color associations for the 199 words tested by \citet{murthy_shades_2022}, for a total of 150 simulated ``subjects'' (i.e., 150 unique prompting contexts for the relevant experiments). 

\paragraph{Diversity metric.} 
In order to examine conceptual diversity in any domain, including color-associations, we need a diversity metric over the space. Here, following \citet{murthy_shades_2022}, we measure the variability between two colors (denoted by $\Delta E$) by their perceptual similarity in CIELAB space. For a given word $w$, we compute internal and population measures of variability as follows.
$\Delta E_{internal} (w)$ measures the $\Delta E$ between the colors chosen by a participant in the first and second blocks for word $w$. $\Delta E_{population} (w)$ captures the similarity between different individuals' color responses for word $w$ (i.e., the average pairwise similarity among every pair of color responses provided by different participants).

For each word $w \in W$, the heterogeneity of a population (i.e., a set of subjects $S$) is then captured by the following metric:
\begin{equation}
    d_w = \frac{1}{S} \sum_{s \in S} \frac{|\Delta E_{internal}(s) - \Delta E_{population}(s)|}{\sqrt{2}} \label{eq:colors-metric}
\end{equation}

Intuitively, this metric measures the divergence from the ``line of unity'' -- i.e., the signature of a homogeneous population in which individuals share the same underlying distributions. The greater this metric, the more conceptual diversity, since the variability between different individuals' distributions is greater than the variability of the individual's own internal distribution ( \Cref{fig:overview}, top).\footnote{The reverse scenario, where $\Delta E_{internal}>\Delta E_{population}$ is also possible, though highly unlikely, especially as the sample size grows. Because the different participants' associations are independent samples, the probability of sampling similar colors (low $\Delta E_{population}$) from different individuals' highly variable internal probability distributions (high $\Delta E_{internal}$) is low.}

\paragraph{Human baseline.}

The relationship between $\Delta E_{internal}$ and $\Delta E_{population}$ for human participants is shown in \Cref{fig:baselines}a, with data reproduced from \citet{murthy_shades_2022}. While these two measures are correlated for humans, internal variability is overall lower than population variability, consistent with a heterogeneous population. For the human data, the average of the metric in \Cref{eq:colors-metric} = 15.82 (95\% CI: [15.18, 16.58]). 

\subsection{Domain 2: Conceptual similarity judgements} \label{sec:concepts-domain}

The second domain, inspired by \citet{marti_latent_2023}, probes the number of latent concept clusters in a population. 
In this task, participants were asked for similarity judgments between concepts in a particular category (e.g., ``Which is more similar to a finch, a whale or a penguin?''). Two categories were tested: common animals, which are more likely to have shared conceptual representations across individuals; and United States politicians, which are likely to have varying representations across individuals.

\paragraph{Model evaluation.}

To measure similarity judgments within concept categories, we presented models with the following query: 
\begin{quote}
    Question: Which is more similar to a \texttt{[TARGET]}, \texttt{[CHOICE1]} or \texttt{[CHOICE2]}? Respond only with ``\texttt{[CHOICE1]}'' or ``\texttt{[CHOICE2]}''.
\end{quote}
\texttt{[TARGET]} is one of the 10 words from the relevant category (animals or politicians), and \texttt{[CHOICE1]} and \texttt{[CHOICE2]} cover all combinations of the remaining words therein. In most cases, model output only one of the two answer choices, but for outputs where additional text was generated, the first sentence of the generation was selected (as it usually contained the model's response) and string matching was used to identify the model's choice. If neither choice was present, the model's response for that query was excluded from further analyses.

\paragraph{Diversity metric.} 
Here, the heterogeneity or diversity of the population intuitively corresponds to the number of latent concepts for each word among individuals in the population. To measure this formally, the similarity judgment for a given word was coded as a binary vector indicating the similarity rating between every other pair of items. Following \citet{marti_latent_2023}, the responses were then analyzed using a non-parameteric Bayesian clustering model (Chinese restaurant process; CRP) to model a distribution over the number of clusters (concepts) for each word. For our analyses, we implement the version of this model that implements a prior preference for fewer clusters and runs inference using a Gibbs sampler.

For each word, the posterior distribution fitted by the CRP was then used to estimate the probability of only a single conceptual representation exists in the population of sampled individuals. 
The higher this probability is, the less diverse the population is. Therefore, as our measure of population heterogeneity in this domain, we compute $1 - P(\text{one concept})$, or equivalently, $P(\text{multiple concepts})$.

\paragraph{Human baseline.}

\citet{marti_latent_2023} find that words in the ``politicians'' category are far more likely to have multiple meanings than ``animal'' words (politicians: 0.69  (95\% CI: [0.66, 0.72]); animals: 0.43 (95\% CI: [0.39, 0.48]); see \Cref{fig:baselines}b). This result reflects the conceptual diversity that is inherent to human populations even in the simplest domains: there exist multiple latent meanings and representations even for words with very concrete, real-world referents. 
Additionally, they find that average between-subject reliability across all concepts is 50\%, meaning that two people picked at random are equally likely to agree or disagree for a given concept.

\subsection{Simulating populations within models} \label{manipulations}

Our main question is whether models can capture human-like conceptual diversity, for the purposes of simulating the response distributions of unique human subjects. 
To investigate this question, we tested two ways of manipulating the ``diversity'' of responses generated by models: increasing the softmax temperature parameter, and conditioning on different types of contexts. These techniques were chosen because of their popularity in related literature, and their accessibility as inference-time interventions for  downstream users. 

\paragraph{Method 1: Increasing temperature.} Increasing the temperature in the softmax normalization during inference-time decoding effectively increases the entropy of the distribution over tokens. Prior work has explored this method as a way of increasing the diversity of generated output \citep{yu_large_2023,tevet_evaluating_2021,peeperkorn_is_2024,bellemare-pepin_divergent_2024}. Most relevant to our work, \citet{wang_large_2024} explore different temperature values as a way to overcome models' flattening of the multi-faceted nature of demographic groups. This work finds that temperature settings that do not just induce diversity in the form of incoherence ($t \in [1.0, 1.2, 1.4]$), also do not do enough to induce meaningful heterogeneity to solve the flattening of demographic groups. 

For completeness and to see how much we can push our measure of diversity, we test some higher values of temperature settings for each model ($t \in [t_0, 1.5, 2.0]$, where $t_0$ is the default temperature of the base model\footnote{Mistral has a default temperature of 1.0, Llama a default of 0.9, and Gemma a default of 0.7.}). 

\paragraph{Method 2: Prompting.} We also tested the effect of ``persona prompting'', where the prompt explicitly instructs the model to simulate a particular individual or group of individuals, usually using group-specific demographic information. 
For example, \citet{wang_large_2024} use persona prompts of the form ``Speak from the perspective of [identity] living in America'', and \citet{he-yueya_psychometric_2024} use more complex prompts such as ``Pretend that you are an 11-year-old student. Your gender is female. You are eligible for free school meals due to being financially disadvantaged.'' While both studies found that such identity-based prompts do not capture relevant metrics to the extent seen in human populations, other work has found success with prompts that simulate expertise and developmental stages \cite{salewski_incontext_2024}.

For our experiments, we adopt the demographic-prompting paradigm, as this approach most closely mimics the process of gathering a representative sample of adult populations for behavioral experiments in our domains. We prepended persona contexts of the following form to the evaluation query:
\begin{quote}
    You are a \texttt{[race]} \texttt{[gender]} from \texttt{[hometown]} in \texttt{[state]} who is \texttt{[age]} and works as a \texttt{[occupation]}.
\end{quote}
We derive the template and field values for this condition from \cite{wang_large_2024} with additions made to the \texttt{[race]} field to cover all the United States census data categories and to the \texttt{[occupation]} field, to induce more diversity across our personas. We populate each field in the context by randomly selecting a value for each.   

We additionally considered three control conditions: \emph{none}, where no context was prepended to the query (matching the prompting setup from the temperature-based manipulations described above); \emph{random}, where we prepended contexts matched in length to the persona-based contexts but on unrelated topics; and \emph{nonsense}, where we scrambled these unrelated contexts at the word-level. The latter two conditions were designed to test whether adding randomness in the form of conditioning models' outputs on highly surprising tokens could sufficiently simulate meaningful human diversity.

To create the \emph{random} and \emph{nonsense} prompt contexts, we first filtered English Wikipedia to remove articles about people (to avoid reduplicating persona prompting) as well as disambiguation pages. We obtained the random contexts by taking the first sentence of a randomly sampled article from the filtered dataset. We obtained the nonsense contexts by shuffling the words within each, and re-applying sentence casing to each resulting bag of words. 

This gave us a total of four prompting conditions, for which the number of prompts varied depending on the domain so as to most faithfully replicate the participant sampling procedures for each experiment ($N_{color}=150, N_{concept}=1800$). In this setting, we used each model's default temperature. 

\begin{figure*}[h!]
  \centering
  \includegraphics[width=.8\textwidth]{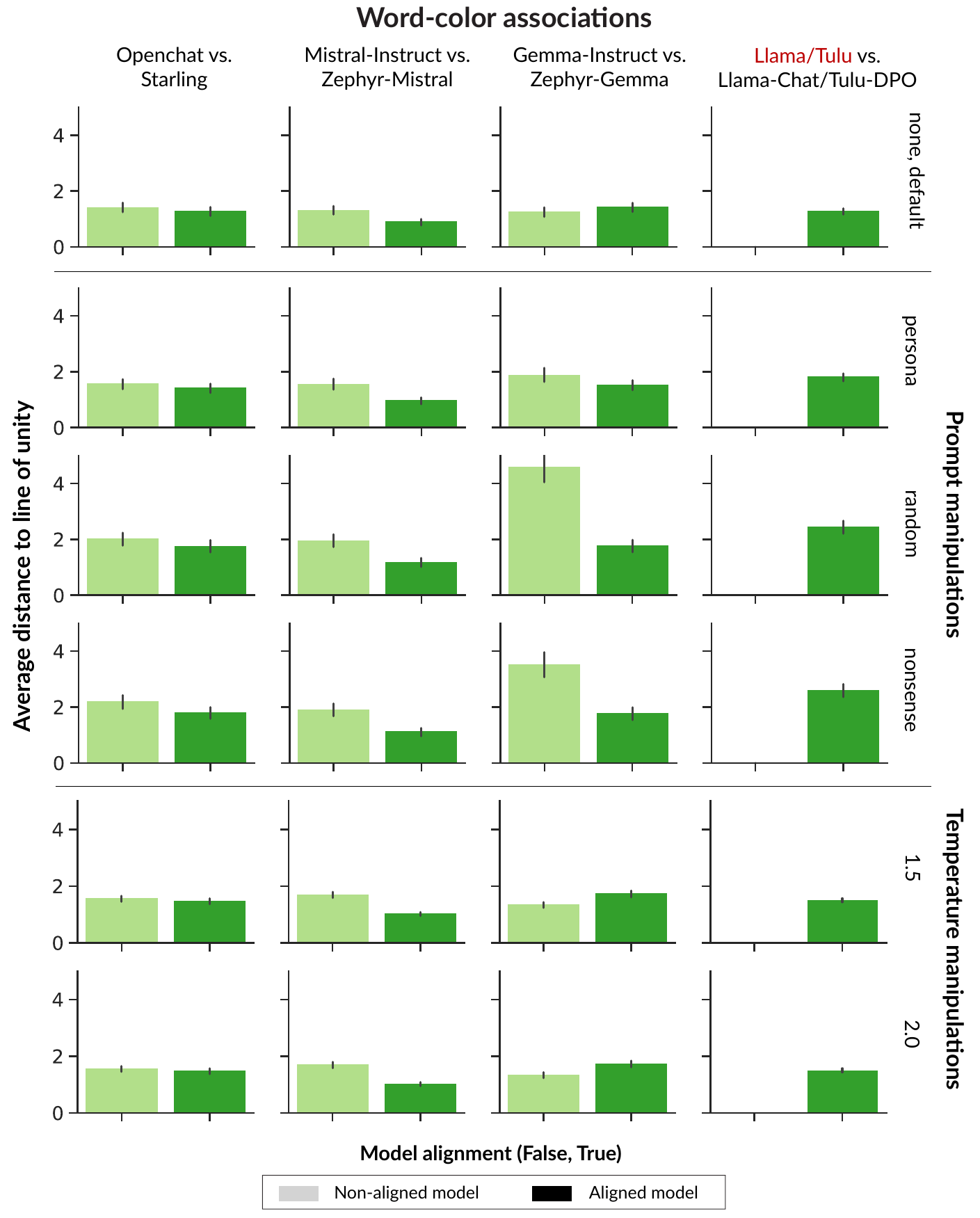}
  \caption{Heterogeneity of simulated LLM population in word-color association domain for prompting and temperature manipulations. The y-axis indicates $d_w$ (\Cref{eq:colors-metric}) averaged over words.
  Rows $=$ baseline, followed by prompting and temperature conditions; columns $=$ model families. Darker bars indicate aligned models. For reference, the human baseline \citep{murthy_shades_2022} is 15.82.}
  \label{fig:colors-results}
\end{figure*}

\begin{figure*}[h!]
  \centering
    \includegraphics[width=\textwidth]{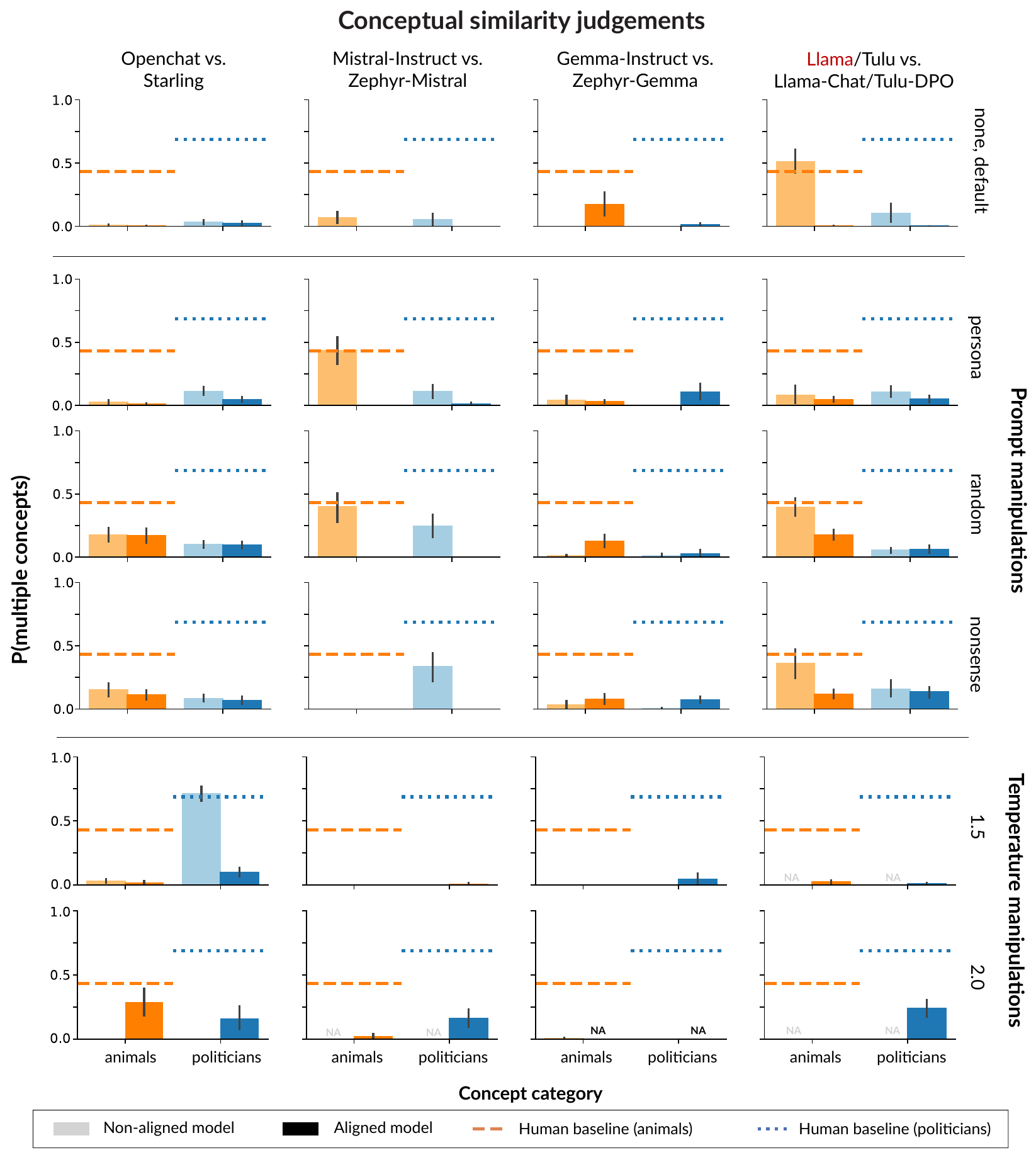}
  \caption{Heterogeneity of simulated LLM population in conceptual similarity domain. The y-axis indicates the probability of more than one conceptual representation, estimated using Chinese Restaurant Process model.
  Rows $=$ baseline, followed by prompting and temperature conditions; columns $=$ model families. Darker bars indicate aligned models. Human baselines for each conceptual category \citep{marti_latent_2023} are shown as horizontal lines.}
  \label{fig:p(1 concept)}
\end{figure*}

\subsection{Models}
We test five pairs of open-source non-aligned and aligned model variants, spanning across three families of base models: Mistral \citep{jiang_mistral_2023}, Gemma \citep{gemma_2024}, and Llama \citep{touvron_llama_2023}. We generally use pairs such that the non-aligned and aligned variants are both fine-tuned on the same base model. This allows us to factor out the effects of the underlying architecture and pre-training, focusing only on the contrast between alignment and other forms of fine-tuning.

Our tested models are summarized in \Cref{tab:models}. Within the Mistral base family, we test two pairs of non-aligned and aligned models: 
Openchat \citep{wang_openchat_2024} vs Starling \citep{zhu_starling-7b_2024}, and Mistral-Instruct \citep{jiang_mistral_2023} vs Zephyr-Mistral \citep{tunstall_zephyr_2024}. For the Gemma base family, we test Gemma-Instruct \citep{gemma_2024} vs Zephyr-Gemma \citep{tunstall_zephyr_2024}. And for the Llama base family, we test two pairs of aligned and non-aligned models: Llama (base) vs Llama-Chat \citep{touvron_llama_2023}, and Tulu vs Tulu-DPO \citep{ivison_camels_2023}.

There are a few exceptions to the general grouping pattern described above. First, Starling undergoes alignment on top of the already-reinforcement-learning-fine-tuned (RLFT) Openchat model, instead of the shared base Mistral model.\footnote{Because these models are still separated by one degree of alignment, like the other model pairs we test, we still refer to this pair as non-aligned vs. aligned in our results for consistency.}
And second, Llama-Chat combines both instruction fine-tuning and alignment into the same model, leaving its non-aligned counterpart, Llama, as the only model in our suite that has also \emph{not} been instruction fine-tuned after pre-training. 

Our tested models span across two popular alignment methods: Reinforcement Learning from Human Feedback \citep[\textbf{RLHF};][]{bai_training_2022,ouyang_training_2022}, and Reinforcement Learning from AI Feedback \citep[\textbf{RLAIF};][]{bai_constitutional_2022}. RLHF involves collecting preference data from human crowdworkers, which is then used either to derive a reward model for subsequent fine-tuning (as in PPO, \citealt{schulman_proximal_2017}; or APA, \citealt{zhu_fine-tuning_2023}) or for direct, reward-free fine-tuning \citep[as in DPO;][]{rafailov_direct_2023}. 
RLAIF involves a similar process, but models are aligned to synthetically generated data instead of human preferences.

In order to isolate the effect of alignment/training methods, we control for model size, testing the 7B-parameter version of all models. All models were downloaded from HuggingFace.

\section{Results}

\subsection{Word-color associations}

\Cref{fig:colors-results} shows the main results from the word-color association domain.
The y-axis shows the heterogeneity metric $d_w$ (\Cref{eq:colors-metric}) averaged over all tested words. We omit the results from two models due to a high number of invalid responses among their generations: Llama and Tulu (see Appendix, \Cref{fig:response_counts} for counts of invalid responses). Overall, we find that none of the models reach the heterogeneity of the human population, with low values for the diversity metric detailed in \Cref{eq:colors-metric} compared to the human baseline of 15.82.
To more closely assess the effects of temperature and prompting manipulations on models' conceptual diversity we constructed a regression model predicting our diversity metric from these conditions with random effects of LLM. In a combined model including both predictors we find a main effect of both prompt and temperature manipulation ($p < 0.001$ for both), but a much higher coefficient for prompts $(b=0.294)$ than for temperatures $(b=0.093)$, with minimal between-group variability among individual LLMs $(b=0.177, SE=0.058)$. These findings suggest that adding noise through prompting techniques is more effective in increasing population heterogeneity than adding noise via softmax temperature.


Turning to the effect of alignment (i.e., comparing light and dark bars within each facet), we find that the aligned model variants have similar or \emph{lower} heterogeneity than their non-aligned counterparts, with the exception of the Gemma models under the temperature manipulations. This is reflected in the results of a regression model predicting our diversity metric from the absence or presence of model alignment with random effects of model family (columns in Figure \ref{fig:colors-results}). We find a significant negative main effect of alignment $(b=-0.495, p < 0.001)$, particularly contributing to a decrease in our diversity metric in this domain, with minimal between-group variability among model families $(b=0.114, SE=0.055)$.

\begin{figure}[t]
  \includegraphics[width=.48\textwidth]{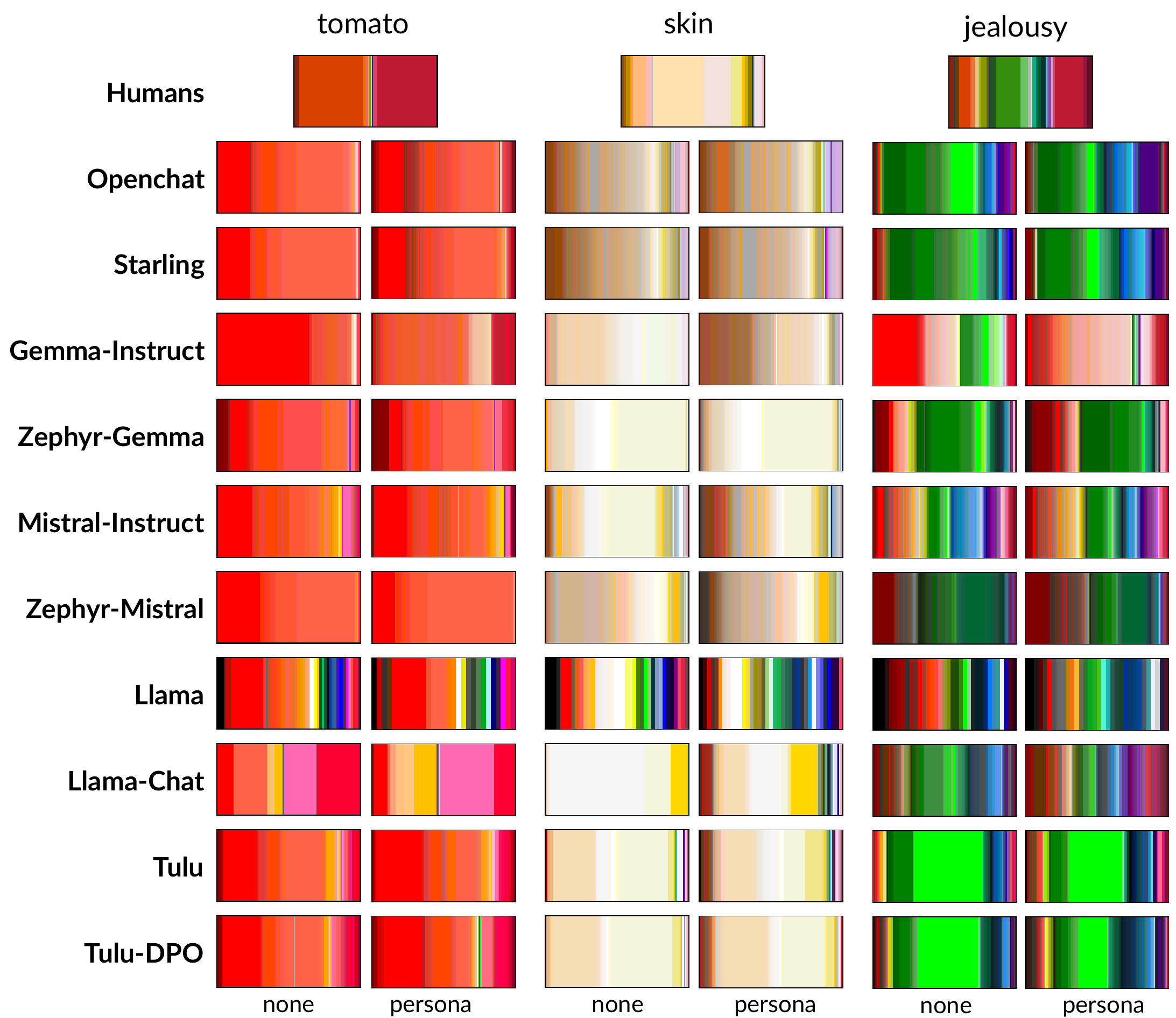}
  \caption{Models' and humans' color associations for a variety of word types. For words where there is high human variation (e.g. abstract words like “optimism”) the models' responses appear to collapse to just handful of colors, but these colors are easily interpretable (e.g. \textit{jealousy}$\approx$green) 
 }
  \label{fig:color_bars}
\end{figure}

We also qualitatively analyzed the similarity between models' and humans' word-color associations. \Cref{fig:color_bars} shows human and model distributions over color space for three example words and the \emph{persona} prompting manipulation relative to the no-prompt baseline. These are well-aligned in some cases, like ``tomato'', which is strongly associated with red. Interestingly, for more abstract words like ``optimism'' or ``fame'' (not pictured), humans have heterogeneous color distributions, whereas models' responses are less diverse but still interpretable -- for example, ``fame'' is associated with gold and ``jealousy'' with green. For the word ``skin'', we see humans' color associations feature the lighter skin tones that one might expect from a Western population. In contrast, some of the models (e.g. Openchat, Starling, and persona-prompted Gemma-Instruct) actually do a better job of diversifying these associations. 
This might suggest that models' word-color associations largely rely on distributional semantics (e.g., common co-occurrences such as ``green with jealousy''), whereas humans might have other associations with abstract concepts that aren't commonly expressed in text corpora. The similarity between models' and humans' color associations also corroborates prior work showing that text-only models can learn structure in perceptual domains such as color space \citep[e.g.,][]{abdou_can_2021,patel_mapping_2022}.

\subsection{Conceptual similarity judgments}

\Cref{fig:p(1 concept)} shows the main results from the conceptual similarity domain. We omit the results from one model in this domain due to a high number of invalid responses among their generations: Llama (see Appendix, \Cref{fig:response_counts} for counts of invalid responses)\footnote{Specific models and manipulations (e.g. high temperature settings) with a high number of invalid responses were also removed post-clustering due to insufficient data and are indicated by ``NA''.}. 

We find that in this domain as well, models largely fall short of the heterogeneity of the human population (horizontal lines), with only a few non-aligned models reaching the human baseline. Across models and concept words, the CRP estimates a very low probability that multiple conceptual representations exist among the population.
Qualitatively, in Figure \ref{fig:p(1 concept)}, we see that with the exception of the Gemma-Instruct and Zephyr-Gemma pair (and in some of the temperature manipulation settings), this effect is more pronounced among the models that have been aligned. The results of a regression model show a significant main effect of alignment $(b=-0.079, p < 0.001)$ on the probability of multiple concepts existing among an LLM-simulated population, with minimal variability across model families $(b=0.002, SE=0.009)$.


The results of a combined regression model predicting our diversity metric in this domain from both predictors of prompt and temperature manipulations (with random effects of model family) showed small, though significant effects of both manipulations $(b=0.019, p<0.001)$ and $(b=-0.018, p<0.001)$, respectively. This result corroborates that of the word-color domain, where prompting manipulations seem to have a greater impact on increasing models' conceptual diversity than temperature manipulations, though the effects are small.


Interestingly, the models also fail to capture another signature pattern in the human data: increased diversity for politician words than animal words (cf.~\Cref{fig:baselines}b). The heterogeneity measure generally stays the same or even slightly decreases between animals and politicians (with a few exceptions). It is reasonable to expect that text-only models would be able to learn conceptual similarity relationships based on the contexts in which the terms appear, and yet, we do not find greater diversity for politician words.
We speculate that this could be related to political concepts being more important ``targets'' for alignment, as they may be associated with more extreme or emotionally charged language.

We also consider a second measure of population heterogeneity: the probability that two random subjects will share similarity judgments for a given concept. The average between-subjects reliability for the human baseline is 50\% (ranging from 33\% to 62\% with no significant differences between animals or politicians). Across temperature and prompt manipulations, models achieve similar or lower average reliability, suggesting high variation between simulated ``subjects''' responses (see Appendix, \Cref{fig:reliability}). However, taken together with the generally low probability of multiple concept clusters existing within these LLM-simulated populations, suggests that this variability does not correspond to meaningful, structured differences in conceptual representations as it appears to in humans.

\section{Discussion}

Our findings corroborate those of recent work in many domains that suggest that injecting variability in models' outputs through temperature and prompt manipulations does not do enough to induce meaningful variability on relevant metrics, and often just leads to highly incoherent outputs.
One reason for this could be that increasing the entropy of the output distribution or conditioning on highly surprising tokens will increase models' uncertainty over subsequent tokens in a domain-general sense. 
However, the kind of uncertainty that gives rise to meaningful individual differences in humans is likely much more constrained to a particular task and cognitive domain than this. 
Exploring more task-specific methods for injecting \emph{structured} randomness into models' generations could help LLMs to overcome their population-average behavior in ways that more meaningfully simulate the cognitive differences in human individuals. 

This raises further questions of what factors give rise to individual differences in humans and whether these are the same attributes that can result in individual-like models. While it is common to obtain a representative samples of a human population for studies in cognitive science using identity attributes like gender or race, the identity-level representations of people that are most easily replicated in post-training methods likely do not capture the reasons for the differences in people's conceptual representations. Exploring corresponding notions of context-specific cognitive or computational resources (e.g., working memory) in models could allow them to serve as testbeds for theories of individual differences in humans \citep[cf.][]{hu2024auxiliary}.

\section{Conclusion}

Here, we used a new approach to evaluate conceptual diversity in synthetically-generated LLM ``populations''. Specifically, we adopted metrics from prior studies of conceptual diversity in humans that relate the variability in individuals' underlying representations to population-level variability. 
We evaluate two popular methods for eliciting unique individuals from LLMs -- temperature and prompting manipulations -- in the word-color and word-concept domains and find that adding noise in the form of unrelated and randomly shuffled prompts does more to increase model diversity than persona prompting and temperature manipulations. 
Across two domains, no model reached human-level conceptual diversity.
Further, our results suggest that alignment to synthetic or human preferences (in the form of RLAIF and RLHF) flattens models' conceptual diversity in these domains, compared to their base or instruction fine-tuned counterparts. 

While alignment has now become a central part of LLM development, our findings urge caution that more work should be done to understand how it affects models' internal diversity of ``opinions'' or conceptual representations. This is especially important given recent calls to use LLMs as stand-ins for human subjects in behavioral research. While it is clear that there could be gains under this paradigm, we should also ask: what might we be losing?

\section*{Limitations}

We tested a limited set of models with a relatively small number of parameters (7B), due to computational resource constraints. In addition, we only tested two popular alignment methods, RLHF and RLAIF. It remains unknown whether larger models, possibly aligned using different techniques or preference datasets, would exhibit the same patterns found in our experiments.

We only tested models in English-language domains, and we compared models to data collected from human participants based in the United States \citep{murthy_shades_2022} or collected by US-based researchers \citep{marti_latent_2023}.
While even these relatively homogeneous participants already exhibit individual-level variability, which LLMs fail to capture, an important consideration for future work is to examine conceptual representations across individuals with more diverse cultural and personal experiences.

Finally, we note that the models were tested in relatively simple domains (word-color associations, and similarity judgments between animals or politicians), which might not reflect the settings or applications in which models will be used.

\section*{Acknowledgments}
We thank the members of the Harvard Computation, Cognition, and Development Lab, as well as Ilia Sucholutsky, for their helpful comments and discussion. We also thank Bala Desinghu for support with computing resources. TU was supported by the Jacobs Foundation. This material is based upon work supported by the NSF Graduate Research Fellowship under Grant No. DGE 2140743 to SKM, and by the Kempner Institute for the Study of Natural and Artificial Intelligence at Harvard University. 
\\

\noindent The code and data for our analyses is available at \href{https://github.com/skmur/onefish-twofish}{ https://github.com/skmur/onefish-twofish}.

\bibliography{custom}

\newpage
\appendix

\section{Percentage of invalid responses}
\label{sec:appendix-valid-responses}

\begin{figure*}[ht]
  \centering
  \includegraphics[width=.8\textwidth]{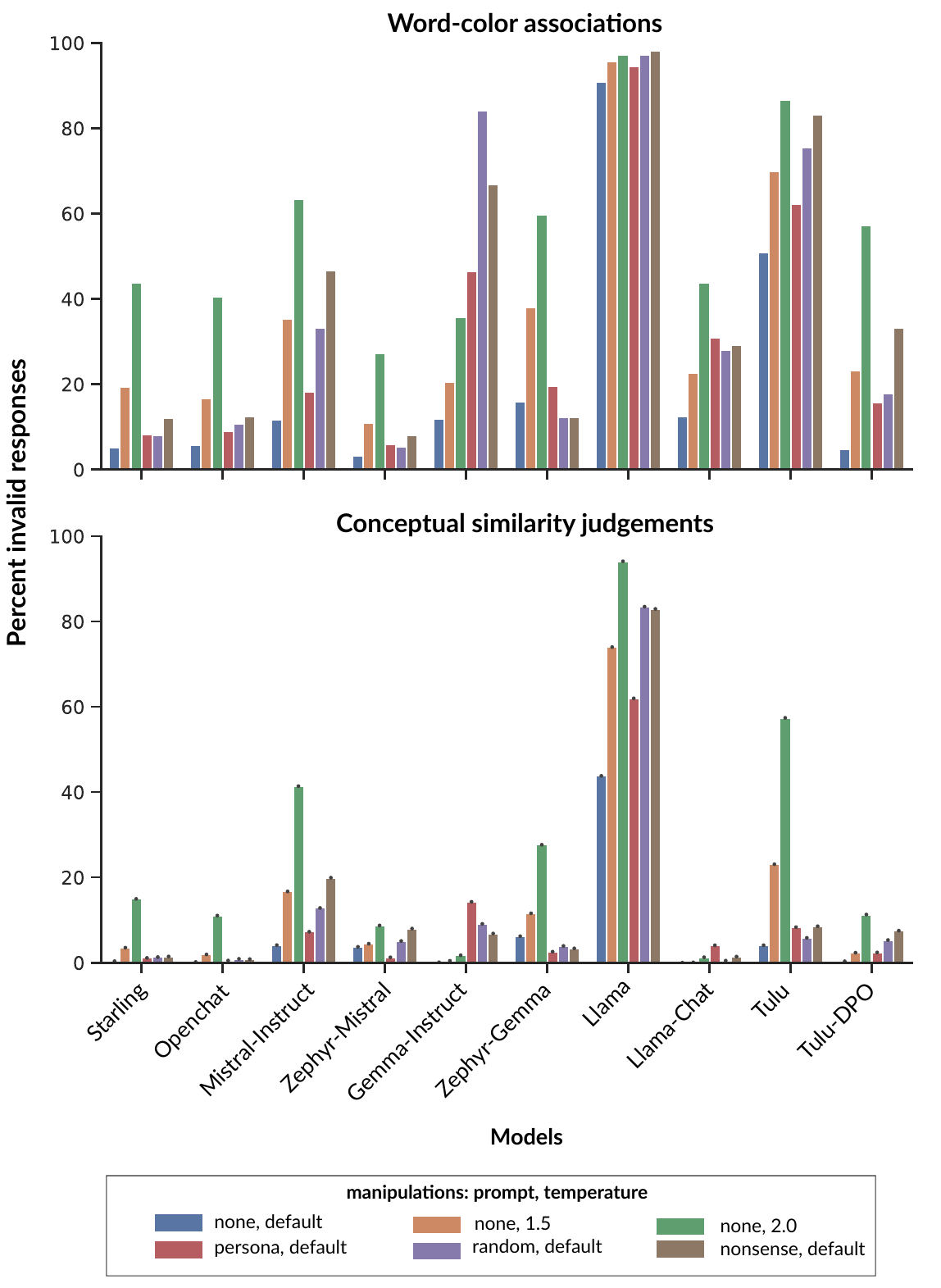}
  \caption{Number of invalid responses for the word-color associations task (top) and conceptual similarity judgements task (bottom) for each tested combination of prompting condition and temperature manipulation. We omit the results of Llama and Tulu on the color task, and for Llama on the concept task given the low  average valid responses across prompt and temperature manipulations. In general, we find that adding noise via high temperature and unrelated prompt contexts increases incoherence, not meaningful diversity.}
  \label{fig:response_counts}
\end{figure*}

\Cref{fig:response_counts} shows the percentage of total queried responses that constituted invalid responses for each model in each domain. We omit the results of models that have an average of 70\% or more invalid responses across all prompt and temperature manipulations: Llama and Tulu on the color task, and Llama on the concept task. In general, we find that adding noise via high temperature and unrelated prompt contexts increases incoherence and thus invalidity, not meaningful diversity.

\section{Population vs internal $\Delta E$}

\begin{figure*}[ht]
  \centering
  \includegraphics[width=\textwidth]{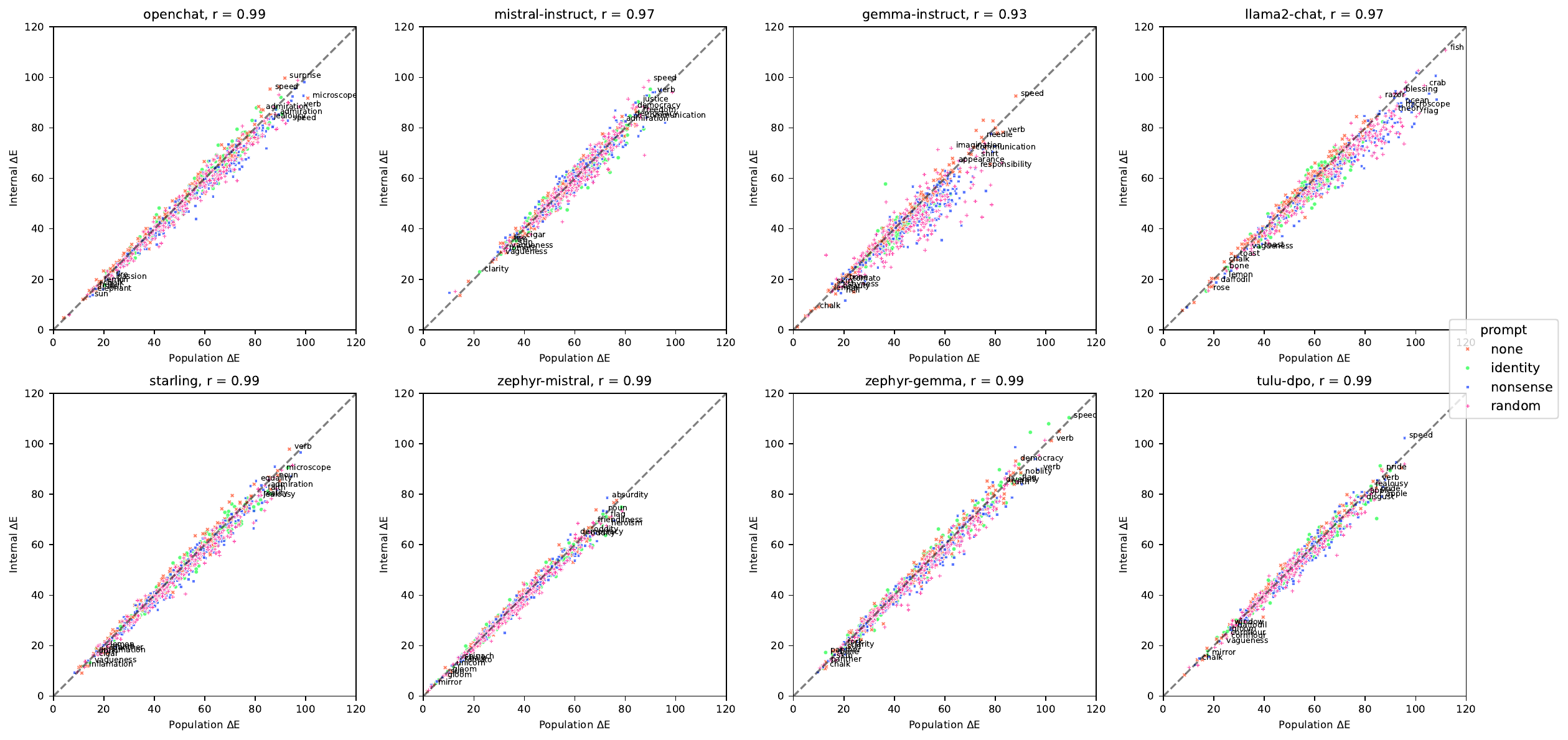}
  \caption{Population vs.~Internal $\Delta E$ for prompting manipulations.}
  \label{fig:models_deltaE_prompt}
\end{figure*}

\begin{figure*}[ht]
  \centering
  \includegraphics[width=\textwidth]{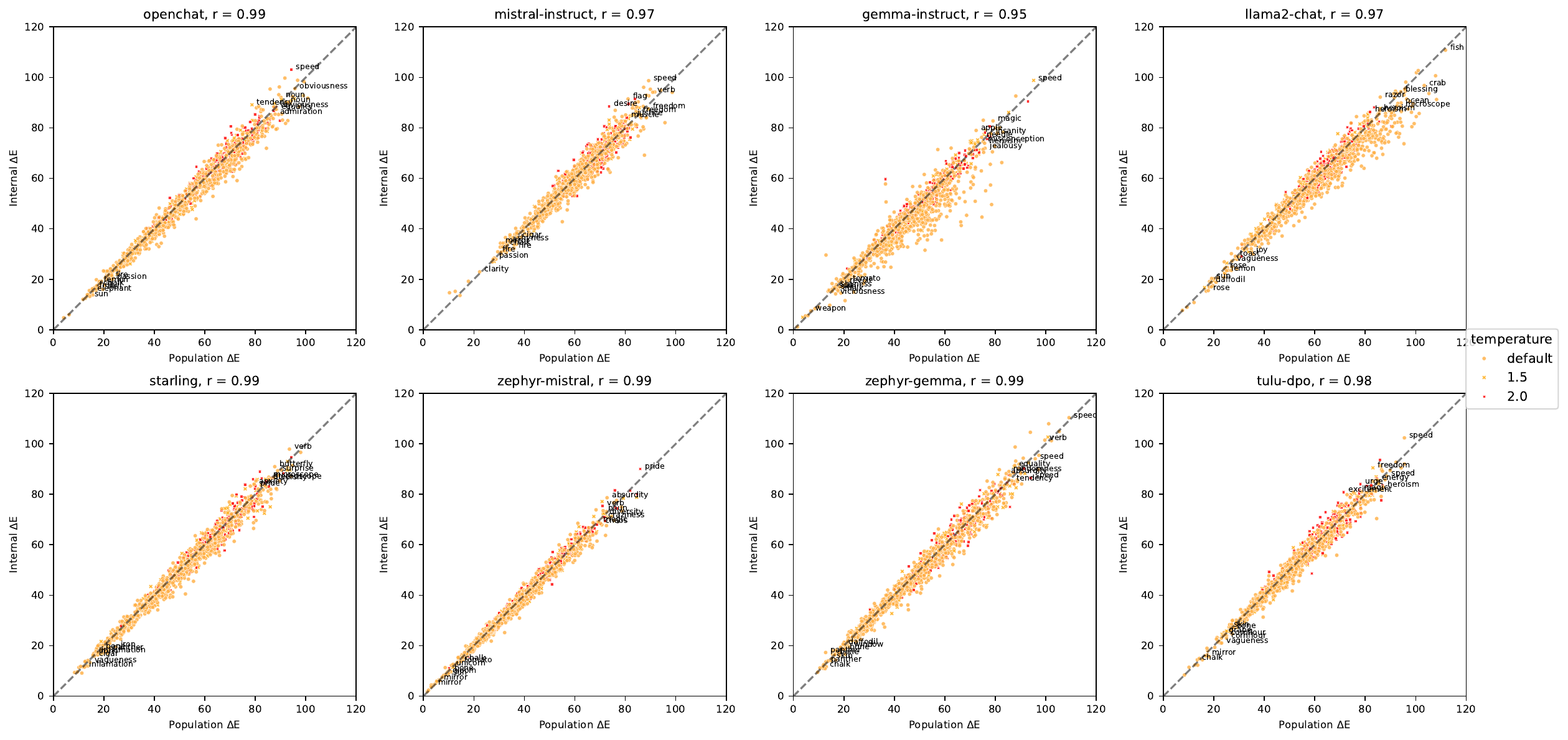}
  \caption{Population vs.~Internal $\Delta E$ for temperature manipulations.}
  \label{fig:models_deltaE_temp}
\end{figure*}

\Cref{fig:models_deltaE_prompt,fig:models_deltaE_temp} show the population $\Delta E$ versus internal $\Delta E$ for prompt- and temperature-based manipulations, respectively. In all settings, the points are near the line of unity, suggesting low population heterogeneity.

\section{Comparing model and human color associations}

\begin{figure*}[ht]
  \centering
  \includegraphics[width=\textwidth]{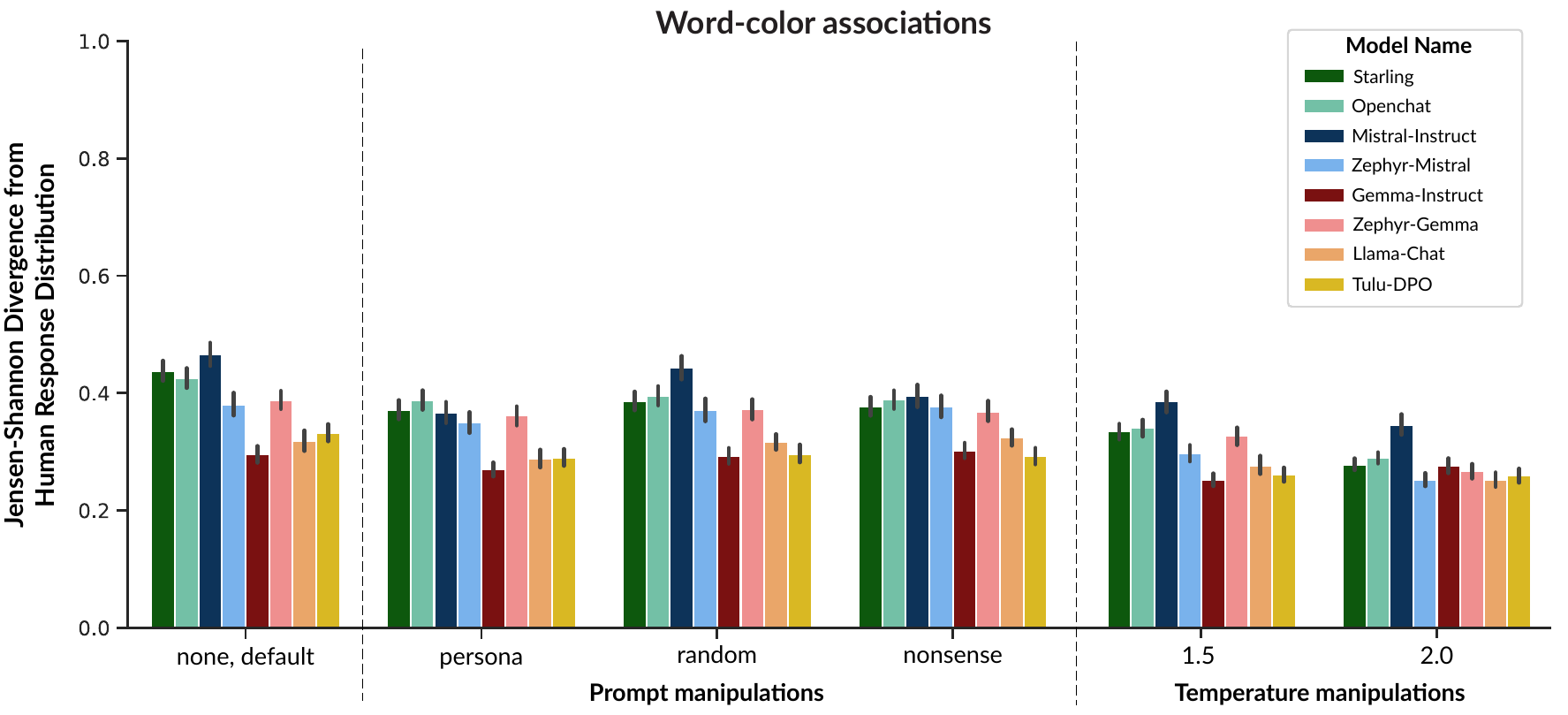}
  \caption{Jensen-Shannon divergence between models’ color response distributions and humans’ for each word. }
  \label{fig:js_divergence}
\end{figure*}

To validate models' ability to perform word-color associations, we report results comparing the similarity between models' and humans' word-color associations. \Cref{fig:js_divergence} shows the Jensen-Shannon divergence between models' and humans' color response distributions, averaged over all words.

\section{Between-subjects reliability for concept task}

As an additional measure of population heterogeneity in the conceptual similarity domain, we calculate the between-subjects reliability for the human baseline and model data (the probability that two random subjects will agree on the similarity judgements for a given concept).

\begin{figure*}[ht]
  \centering
  \includegraphics[width=\textwidth]{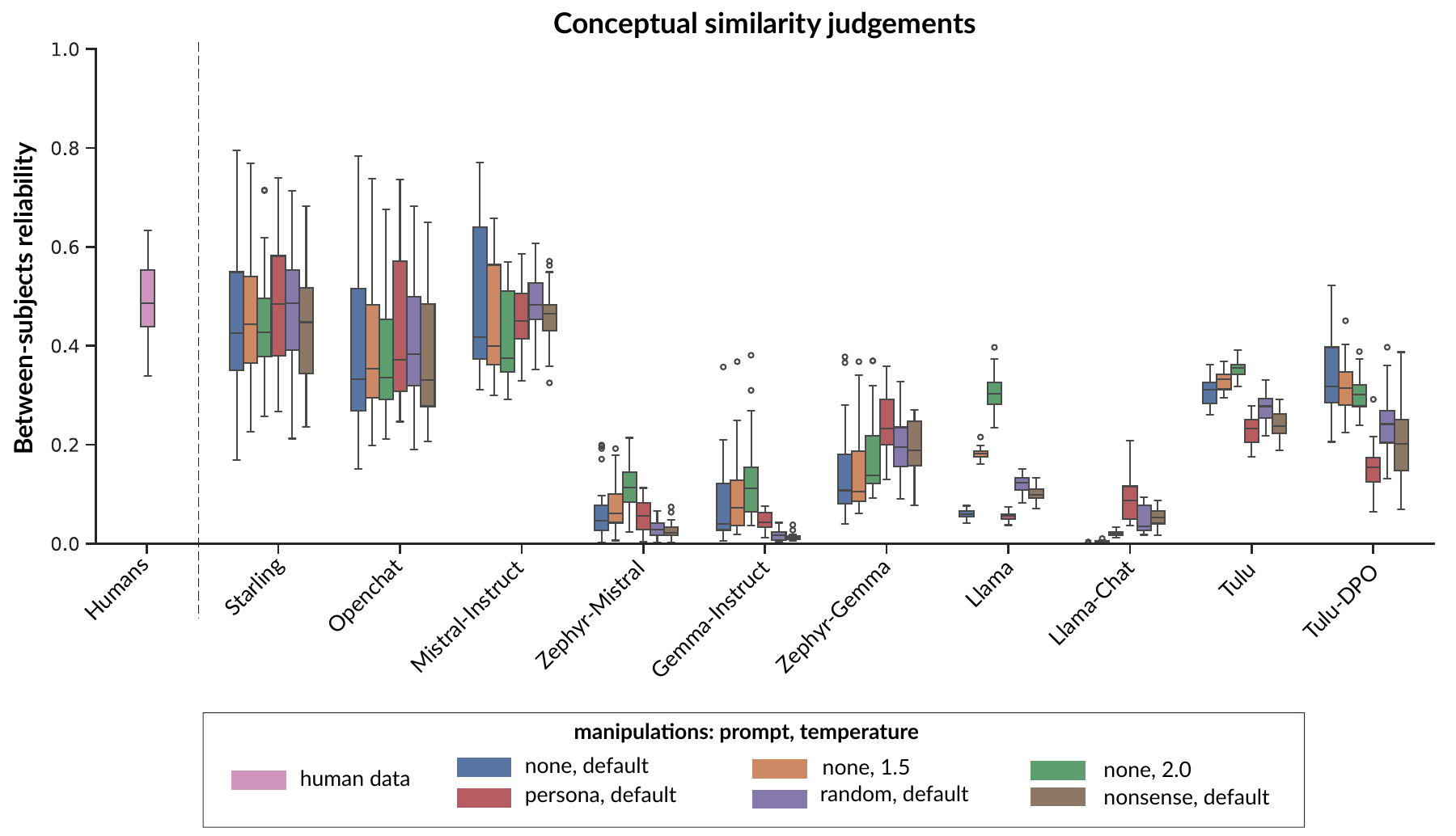}
  \caption{Between-subjects reliability for model data and human baseline in the conceptual similarity domain.
  }
  \label{fig:reliability}
\end{figure*}

\end{document}